\documentclass[conference]{IEEEtran}
\IEEEoverridecommandlockouts
\usepackage{cite}
\usepackage{xurl}
\usepackage{amsmath,amssymb,amsfonts}
\usepackage{algorithmic}
\usepackage{graphicx}
\usepackage{textcomp}
\usepackage{xcolor}
\usepackage{booktabs}
\usepackage{siunitx}
\usepackage{multirow}
\usepackage{makecell}
\usepackage{todonotes}
\setuptodonotes{size=\scriptsize}

\def\BibTeX{{\rm B\kern-.05em{\sc i\kern-.025em b}\kern-.08em
    T\kern-.1667em\lower.7ex\hbox{E}\kern-.125emX}}

\usepackage{tikz}
\newcommand{\ieeecopyright}{%
  \begin{tikzpicture}[remember picture,overlay]
    \node[anchor=south, yshift=0.4in] at (current page.south) {%
      \parbox{\textwidth}{\scriptsize\centering
        Accepted for publication at the IEEE Sensors Applications Symposium (SAS 2026). \\
        \copyright 2026 IEEE. Personal use of this material is permitted. Permission from IEEE must be obtained for all other uses, in any current or future media, including reprinting/republishing this material for advertising or promotional purposes, creating new collective works, for resale or redistribution to servers or lists, or reuse of any copyrighted component of this work in other works.}};%
  \end{tikzpicture}%
}

\begin{document}

\title{
  SAM3-Assisted Training of Lightweight YOLO Models for Precision Pig Farming
  %: A Cost-Benefit Analysis of Foundation Model Annotation
}

\author{
    \IEEEauthorblockN{
        Marcos Vinícius Mendes Faria\IEEEauthorrefmark{1},
        Thiago Borges Pereira\IEEEauthorrefmark{1},
        Isabella C.F.S. Condotta\IEEEauthorrefmark{2},\\
        Thiago Meireles Paixão\IEEEauthorrefmark{1} and
        Francisco de Assis Boldt\IEEEauthorrefmark{1}
    }
    \IEEEauthorblockA{
        \IEEEauthorrefmark{1}\textit{Coordenadoria de Informática}, \textit{Instituto Federal do Espírito Santo (IFES)}, Serra, ES\\
        Email: \{marcos.mendes, thiago.pereira\}@estudante.ifes.edu.br, \{thiago.paixao, franciscoa\}@ifes.edu.br
    }
    \IEEEauthorblockA{
        \IEEEauthorrefmark{2}\textit{Department of Animal Sciences}, \textit{University of Illinois at Urbana-Champaign}, USA\\
        Email: icfsc@illinois.edu
    }
}

\maketitle
\ieeecopyright

% \author{
% \IEEEauthorblockN{1\textsuperscript{st} Marcos Vinícius Mendes Faria}
% \IEEEauthorblockA{\textit{Coordenadoria de Informática}\\
% \textit{Instituto Federal do Espírito Santo}\\
% Serra, ES\\
% marcos.mendes@estudante.ifes.edu.br}
% \and
% \IEEEauthorblockN{2\textsuperscript{nd} Thiago Borges Pereira}
% \IEEEauthorblockA{\textit{Coordenadoria de Informática}\\
% \textit{Instituto Federal do Espírito Santo}\\
% Serra, ES\\
% thiago.pereira@estudante.ifes.edu.br}
% \and
% \IEEEauthorblockN{3\textsuperscript{rd} Isabella C.F.S. Condotta}
% \IEEEauthorblockA{\textit{Department of Animal Sciences}\\
% \textit{University of Illinois}\\
% USA\\
% icfsc@illinois.edu}
% \and
% \IEEEauthorblockN{4\textsuperscript{th} Thiago Meireles Paixão}
% \IEEEauthorblockA{\textit{Coordenadoria de Informática}\\
% \textit{Instituto Federal do Espírito Santo}\\
% Serra, ES\\
% thiago.paixao@ifes.edu.br}
% \and
% \IEEEauthorblockN{5\textsuperscript{th} Francisco de Assis Boldt}
% \IEEEauthorblockA{\textit{Coordenadoria de Informática}\\
% \textit{Instituto Federal do Espírito Santo}\\
% Serra, ES\\
% franciscoa@ifes.edu.br}
% }

\author{
    \IEEEauthorblockN{
        Marcos Vinícius Mendes Faria\IEEEauthorrefmark{1},
        Thiago Borges Pereira\IEEEauthorrefmark{1},
        Isabella C.F.S. Condotta\IEEEauthorrefmark{2},\\
        Thiago Meireles Paixão\IEEEauthorrefmark{1} and
        Francisco de Assis Boldt\IEEEauthorrefmark{1}
    }
    \IEEEauthorblockA{
        \IEEEauthorrefmark{1}\textit{Coordenadoria de Informática}, \textit{Instituto Federal do Espírito Santo (IFES)}, Serra, ES\\
        Email: \{marcos.mendes, thiago.pereira\}@estudante.ifes.edu.br, \{thiago.paixao, franciscoa\}@ifes.edu.br
    }
    \IEEEauthorblockA{
        \IEEEauthorrefmark{2}\textit{Department of Animal Sciences}, \textit{University of Illinois at Urbana-Champaign}, USA\\
        Email: icfsc@illinois.edu
    }
}

\maketitle

\begin{abstract}
Deep learning-based object detection has revolutionized Precision Livestock Farming (PLF), yet a critical barrier remains: high-performance Foundation Models (such as SAM 3) are too computationally intensive for edge deployment, while lightweight models (like YOLO) require prohibitive manual annotation efforts. This work proposes a fully automated knowledge distillation pipeline that leverages the Segment Anything Model 3 (SAM 3) to generate zero-shot pseudo-labels for training efficient YOLOv8 detectors. By treating SAM 3 as an offline auto-annotator, we eliminate the manual labeling bottleneck, producing models capable of real-time inference on resource-constrained hardware. We systematically evaluate this approach on the PigLife dataset, comparing SAM 3-supervised models against human-annotated baselines. Results demonstrate that a SAM 3-trained YOLOv8m achieves a mean Average Precision (mAP) of 79.4\% without human intervention, while reducing inference latency by approximately 200$\times$ compared to the teacher model. Furthermore, stratified analysis reveals that in low-occlusion scenarios, the automated pipeline achieves detection rates comparable to human benchmarks ($AP_{50} > 99\%$). These findings indicate that foundation models can serve as effective, zero-annotation-cost supervisors, enabling scalable edge computing solutions for smart agriculture.
\end{abstract}


\section{Introduction}

The rapid advancement of deep learning-based object detection has revolutionized Precision Livestock Farming (PLF). By integrating continuous sensor data with intelligent processing, PLF enables the automated, real-time management of animal health and welfare \cite{b1, b2, b3}. However, a fundamental tension currently limits the widespread deployment of these systems in smart agriculture. On one hand, emerging foundation models, specifically the most recent Segment Anything Model 3 (SAM 3), released by Meta in November 2025 \cite{b4}, represent a new paradigm in computer vision. Unlike its predecessor SAM 2 \cite{b5}, which relies heavily on spatial prompts such as points or boxes, SAM 3 introduces Promptable Concept Segmentation (PCS), allowing for zero-shot segmentation via semantic inputs without domain-specific training. Yet, these models demand substantial computational resources, rendering them impractical for edge devices in resource-constrained farm environments \cite{b6, b7}. On the other hand, lightweight architectures, such as YOLOv8 nano and small variants, suit embedded hardware but traditionally require large, precisely annotated datasets to achieve competitive accuracy \cite{b8, b9, b10}.

\begin{figure*}[t]
  \centering
  \includegraphics[width=\linewidth]{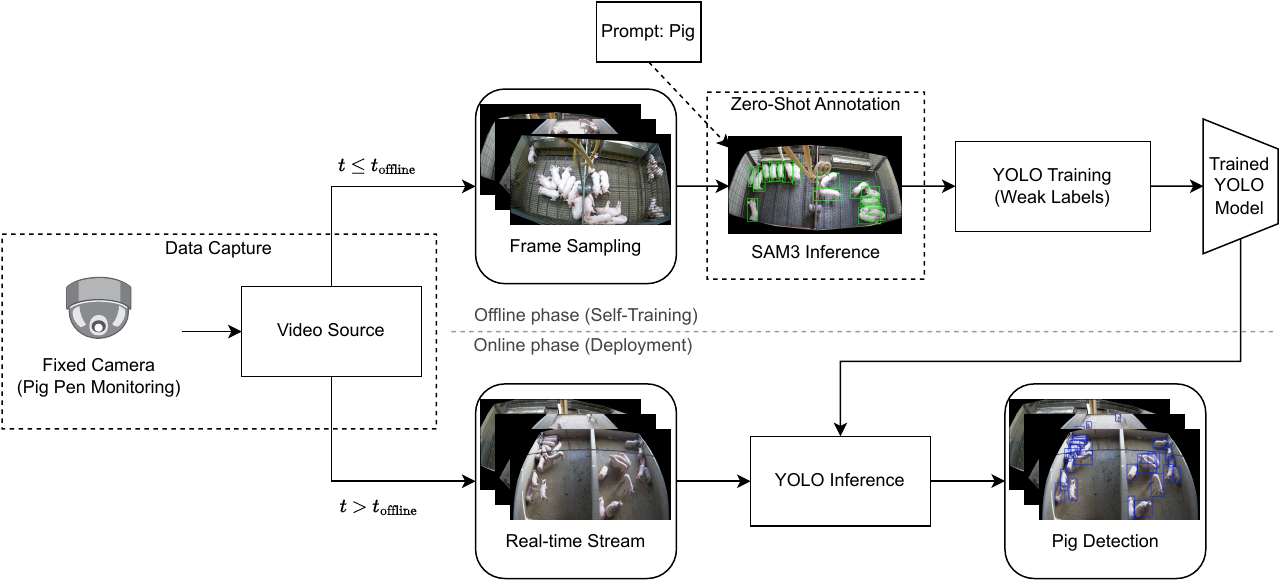}

  \caption{Overview of the SAM-3-assisted pig detection pipeline illustrated in two stages: an offline self-training phase, where the foundation model generates weak labels, and an online inference phase, where the YOLO detector performs real-time inference on live video streams.}

  \label{fig:pipeline}
\end{figure*}

This reliance on manual annotation is a critical bottleneck for producing machine learning models that generalize across domains, including in PLF applications. In realistic farm conditions, where occlusion, crowding, and varying pen layouts create significant domain shifts, scaling solutions across multiple farms requires constant re-annotation \cite{b11}. Previous benchmarking on the PigLife dataset demonstrated that while medium-sized models (YOLOv8m) offer an optimal trade-off between accuracy ($91.8\%$ mAP) and complexity, maintaining this performance across new environments is labor-intensive \cite{b12}. Lee et al. \cite{b13} proposed a ``Single Label On Target'' (SLOT) approach, using genetic algorithms to optimize data augmentation from a single image. While such techniques reduce the volume of required labels, they still depend on initial human inputs and do not fully exploit the semantic reasoning of modern foundation models.

Recent methodologies have increasingly leveraged teacher-student paradigms to automate labeling; for instance, Wutke et al. \cite{b14} introduced a ``Noisy Student'' approach for automatic farrowing monitoring, using pseudo-labels for newborn piglets to increase the F1-score from $0.672$ to $0.922$. Similarly, in pre-clinical imaging, Rickmann et al. \cite{b15} demonstrated that foundation models can serve as reliable pseudo-label generators for porcine cardiac Computed Tomography (CT) segmentation, refining labels through iterative self-training. To address the computational barrier of foundation capabilities, Zeng et al. \cite{b16} presented EfficientSAM3, which employs Progressive Hierarchical Distillation (PHD) to transfer semantic knowledge from SAM 3 to lightweight students. Despite these advancements, the use of SAM 3 PCS to directly facilitate the training of lightweight YOLOv8 detectors for pig monitoring remains underexplored.

This work addresses this gap by exploring the feasibility of SAM 3 as an automatic offline annotator. We investigate the hypothesis that annotations generated by a heavy foundation model can serve as effective supervision for the training of lightweight detectors. Our study analyzes the resulting performance trade-offs to determine whether this approach can significantly reduce human effort while maintaining sufficient accuracy for robust, on-pen deployment. In summary, the main contributions of this paper are:

\begin{itemize}
    \item The introduction of a novel, fully automated detection pipeline for swine monitoring that utilizes SAM 3 as a zero-shot teacher to supervise lightweight YOLO student models.
    \item A systematic evaluation and ablation study quantifying the performance-cost trade-offs across different model scales (nano, small, and medium), elucidating the impact of foundation-model-generated labels on detection accuracy.
    \item A comparative study establishing state-of-the-art results for automated annotation in PLF; specifically, our SAM 3-trained YOLOv8m variant achieves a mean Average Precision (mAP) of $79.4\%$ compared to the human-annotated baseline of $91.7\%$, while requiring zero manual labeling effort.
\end{itemize}

The remainder of this paper is organized as follows: Section II describes the proposed SAM 3-assisted self-training pipeline and application scenario. Section III details the experimental methodology, including dataset characteristics and evaluation metrics. Section IV presents the results and a scenario-stratified discussion. Finally, Section V concludes the work and outlines future research directions.

\section{SAM 3--Assisted Self-Training Pig Detection System}

The architectural overview of the proposed self-training framework is illustrated in Figure \ref{fig:pipeline}. The pipeline operates by dividing the video stream from a fixed camera into two distinct temporal phases: an offline self-training phase and an online inference phase. In the offline stage ($t \le t_{\text{offline}}$), frames are sampled and annotated automatically using the SAM 3 via zero-shot prompting (the text prompt is simply ``Pig''). These weak labels are then utilized to train a lightweight YOLO detector. Finally, in the online phase ($t > t_{\text{offline}}$), the trained lightweight model is deployed for real-time pig detection on edge hardware. The following sections describe each component of the pipeline and the training strategy.

\subsection{Application Scenario}

The proposed system is tailored for continuous operation in realistic PLF environments. The setup utilizes a single fixed camera mounted above the pig pen to monitor animal activity. While the framework is extensible to multi-camera setups, this study focuses on a single-viewpoint constraint to isolate the efficacy of the self-training mechanism. In this context, video data is acquired over a predefined period, from which frames are extracted to construct a training set that accurately reflects the target environment. A critical advantage of this setup is the operational consistency: both training and deployment phases share the same visual conditions, including camera viewpoint, lighting, pen layout, and animal behavior. This alignment ensures that the automatically generated dataset is highly representative of the inference domain, reducing the need for complex domain adaptation.

\subsection{Self-Training Pipeline Overview}

% The core methodology follows a two-stage process, as detailed in Figure \ref{fig:pipeline}. This pipeline is designed to automate the transition from raw video to a deployable edge model without manual intervention.

As illustrated in Figure \ref{fig:pipeline}, the (offline) self-training phase leverages SAM 3 as an automated annotator to generate a dataset with pig bounding-box annotations. Frames sampled from the video source are processed via zero-shot annotation, where the foundation model is conditioned with the text prompt ``Pig'' to locate swine within the pen.
% generic text prompts (e.g., ``Pig'') to identify animals within the pen.
The resulting bounding boxes are obtained directly from the model output to produce weak labels (pseudo-annotations). In this approach, the foundational model plays the role of the human annotator, thus, eliminating the manual labeling effort.

%  This automated process effectively eliminates the manual annotation bottleneck.

The annotated dataset is subsequently used to train a lightweight YOLO detector. Upon completion of training, the system transitions to the online inference phase. In this stage, the computationally intensive foundation model is decoupled from the loop, and the trained lightweight model is deployed for real-time pig detection. This architecture ensures that the system meets the low-latency and resource constraints of edge hardware while leveraging the superior generalization capabilities of SAM 3 for supervision. The next section details the experimental evaluation of the proposed self-trained detection framework.

\section{Experimental Methodology}

\subsection{Dataset}

The experiments utilized the PigLife dataset \cite{b17}, developed by the AIFARMS Institute at the University of Illinois. While the complete dataset encompasses different phases of the swine production cycle, the image-based subset focuses on five specific stages. This subset comprises a total of 2,130 images distributed across gestation (598 images), farrowing (190 images), nursery (600 images), estrus (222 images) and growth (520 images). The raw surveillance footage was curated to exclude human activity and environmental noise, with static frames sampled at 1-second intervals to yield 21,869 individual pig instances.

%{'Nursery': 600, 'Gestation': 598, 'Growth': 520, 'Estrus': 222, 'Farrow': 190}

The original dataset has predefined 80\%-20\% train-test splits. To create a validation split, we further randomly sampled 340 images from the training set, resulting in a final distribution of 1,364 images (65\%) for training, 340 images (15\%) for validation, and 426 images (20\%) for testing. The original annotations in the training split serve two purposes: (i) evaluating the fidelity of SAM 3-generated pseudo-labels and (ii) training baseline models for comparison with our method.

\subsection{Overall Comparative Evaluation}
\label{sec:overall}

Two main research questions guide the overall comparative evaluation:

\begin{itemize}
  \item \textbf{RQ1}: Does the teacher model (SAM~3) have the capability to solve the swine-detection task?
  \item \textbf{RQ2}: Do the lightweight student models obtained with the proposed self-training approach perform similarly to the teacher model?
\end{itemize}

To answer RQ1, we compare the performance of the teacher model (SAM~3) in a zero-shot setting (i.e., without domain-specific fine-tuning) against models trained directly with human-annotated data (baseline models). To answer RQ2, we compare the teacher's performance with that of the student models produced with our self-training approach.

Performance is measured by evaluating the predicted bounding boxes against the expert-annotated ground truth using standard Common Objects in Context (COCO) metrics (\mbox{\url{https://cocodataset.org/#detection-eval}}) (mAP, $AP_{50}$, $AP_{75}$, $AP_{M}$, and $AP_{L}$). The primary metric is Mean Average Precision (mAP) averaged over IoU thresholds from $0.50$ to $0.95$. We also report inference latency (ms) to validate suitability for real-time deployment. 

More details of the evaluation procedure are provided in the following.

\subsubsection{Teacher Evaluation}

This evaluation of SAM~3 (teacher model) on the test set, without domain-specific fine-tuning, provides an expected upper bound for the lightweight student models, since they are trained to mimic the teacher's outputs. We utilized the \texttt{facebook/sam3} checkpoint (\mbox{\url{https://huggingface.co/facebook/sam3}}) in its default configuration to perform zero-shot detection by prompting SAM~3 with ``Pig'' for each test image.

\subsubsection{Students Evaluation}
\label{sec:students}

The student models are produced using our self-training framework, in which SAM~3 acts as an offline annotator. To this end, SAM~3 is used to generate a new set of bounding-box annotations for the 1,704 training and validation images without human intervention. A confidence-score threshold of 0.4 was selected to prioritize recall slightly over precision, ensuring that objects under challenging conditions (occlusion/low light) are not missed while maintaining an acceptable false positive rate. The resulting predictions are projected into absolute pixel coordinates $(x_{min}, y_{min}, w, h)$ and converted into the normalized YOLO annotation format.

The pseudo-labeled dataset is then used to independently train three lightweight YOLOv8 variants sourced from the Ultralytics framework: nano ($n$), small ($s$), and medium ($m$). To ensure a consistent experimental setup, each variant was trained for 100 epochs with a batch size of 4 and a standardized input resolution of $640 \times 640$ pixels. Default hyperparameters provided by the framework—including the SGD optimizer and learning rate schedule—were retained to establish a reproducible baseline. Finally, the trained models were evaluated on the test set and performance metrics were computed.

\subsubsection{Baseline Evaluation}

Training lightweight YOLO models directly with the original training annotations, manually produced by human experts, allows us to establish an overall upper bound for the task. This helps us quantify the performance degradation (in terms of precision) of alternative automatic approaches. To this end, the same YOLOv8 variants (nano, small, and medium) were trained on the original PigLife training set following the configuration described in Section \ref{sec:students} for comparability.

\subsection{Scenario-Stratified Evaluation}
\label{sec:stratified-eval}

\begin{figure*}[tbh!]
  \centering
  \includegraphics[width=\linewidth]{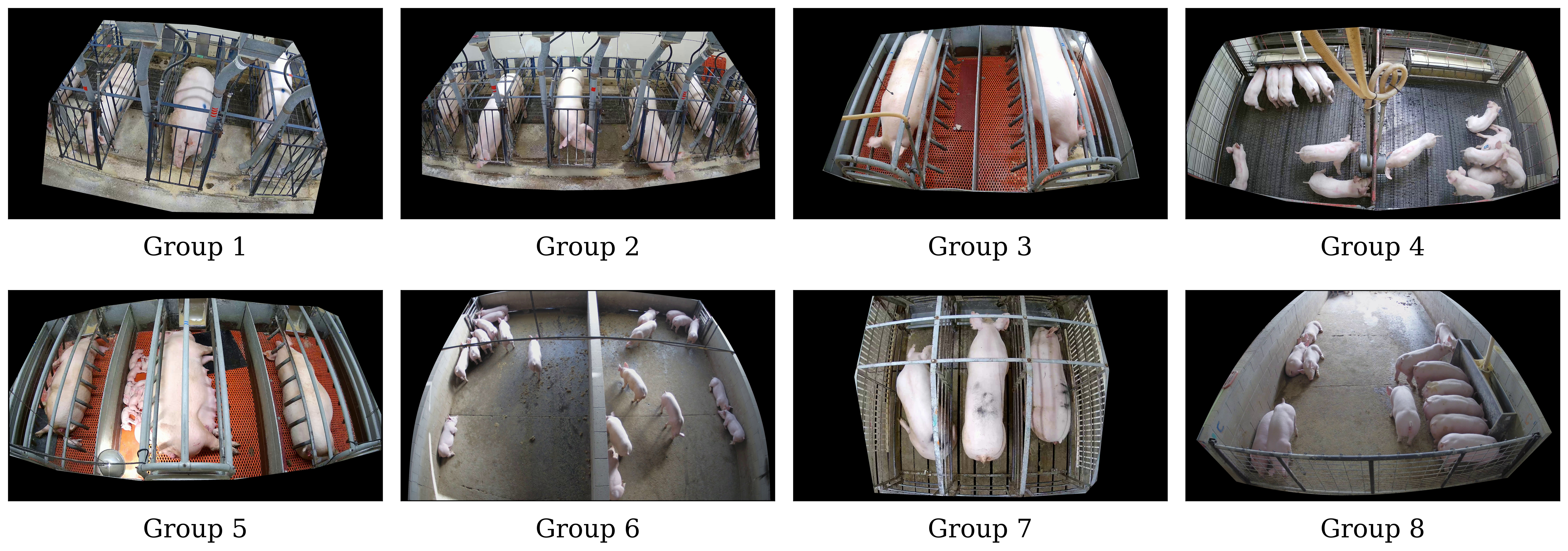}
  \caption{Visual breakdown of the eight test scenarios (Groups 1--8).}
  \label{fig:8-groups}
\end{figure*}

To move beyond aggregate metrics and identify specific failure modes of foundation-model supervision, we conducted a stratified analysis of the test set. Specifically, we manually inspected the test images and grouped them into eight scenarios (Figure~\ref{fig:8-groups}) based on visual characteristics such as camera angle, pen layout, and lighting conditions:

\begin{itemize}
  \item \textbf{Groups 1--2 (Gestation):} Similar settings, but with different camera angles and fields of view. Group~1 focuses on three stalls, whereas Group~2 captures a wider view with five.
  \item \textbf{Group 3 (Gestation):} A distinct pen layout with slatted flooring.
  \item \textbf{Group 4 (Nursery):} A top-down view within a metal pen divided into two sections.
  \item \textbf{Group 5 (Farrowing):} The most challenging setting, with sows and numerous piglets. High density leads to frequent severe occlusions, making accurate annotation and prediction particularly difficult.
  \item \textbf{Group 6 (Growth):} A pen bisected by a central column; includes low-light night imagery.
  \item \textbf{Group 7 (Estrus):} A compact layout separated into three individual stalls.
  \item \textbf{Group 8 (Growth):} An open pen layout without physical dividers.
\end{itemize}

To analyze performance stability across environments, we used the YOLOv8s student model (Section~\ref{sec:students}) as the representative benchmark, selected for its balance between model capacity and computational efficiency.

\subsection{Computational Resources}

All experiments were conducted on a single local PC equipped with an Intel Core i9-10900KF 3.70GHz CPU, 32GB of RAM, and an NVIDIA GeForce RTX 3060 GPU with 12GB of dedicated memory.

\section{Results and Discussion}

\newcommand{\approxm}[1]{$\sim$#1}

\begin{table*}[t]
  \caption{Object detection performance (COCO metrics) of YOLOv8 models. Values in parentheses denote the change after filtering the images.}
  \centering
  \label{tab:yolo_ap_metrics}
  \begin{tabular}{c l c c c l l l l l} 
  \toprule
  \textbf{Annotation} & \textbf{Model} & \textbf{Params (M)} & \textbf{Inf. Forward (ms)} & \textbf{Inf. Pipeline (ms)} & \textbf{$mAP$} & \textbf{$AP_{50}$} & \textbf{$AP_{75}$} & \textbf{$AP_{M}$} & \textbf{$AP_{L}$} \\
  \midrule
  \multirow{3}{*}{\makecell{Human\\ annotated}} 
    & YOLOv8n & 3.2  & 5.70 & 9.29 & 88.1 & 98.9 & 95.7 & 67.5 & 89.0 \\
    & YOLOv8s & 11.2 & 6.10 & 9.51 & 90.6 & 99.0 & 96.9 & 72.1 & 91.4 \\
    & YOLOv8m & 25.9 & 14.70 & 15.43 & 91.7 & 99.0 & 97.8 & 75.2 & 92.5 \\
  \midrule
  \multirow{3}{*}{\makecell{SAM 3\\ generated}} 
    & YOLOv8n & 3.2  & 5.70 & 10.13 & 76.7 & 93.3 & 86.4 & 27.8 & 78.1 \\
    & YOLOv8s & 11.2 & 6.10 & 10.37 & 78.7 & 93.5 & 87.7 & 32.1 & 80.1 \\
    & YOLOv8m & 25.9 & 14.70 & 16.71 & 79.4 & 93.6 & 88.2 & 30.6 & 80.8 \\
  \midrule
  \makecell{Zero-shot\\ baseline} & SAM 3 & \approxm{850} & 1197.18 & 1242.53 & 80.7 & 93.6 & 88.4 & 26.9 & 82.4 \\
  \bottomrule
  \end{tabular}
\end{table*}

% 426.20 ms ± 8.14 ms (Using bfloat16)

\subsection{Overall Comparative Analysis}

To address the research questions presented in Section \ref{sec:overall}, we begin by analyzing the comparative performance of the teacher, student, and baseline models presented in Table~\ref{tab:yolo_ap_metrics}.

Regarding teacher capability (\textbf{RQ1}), the results suggest that SAM 3 demonstrates strong generalization capabilities for this task, even without domain-specific training. The zero-shot teacher achieved an mAP of 80.7 and an $AP_{50}$ of 93.6. While this represents a performance gap of 11 points compared to the human-annotated YOLOv8m baseline (mAP 91.7), the high $AP_{50}$ indicates that the foundation model identifies the majority of animal instances, supporting its potential as a supervisor for automated annotation.

Turning to student-teacher similarity (\textbf{RQ2}), the comparison between the ``SAM 3 generated'' students and the ``Zero-shot baseline'' teacher reveals a strong alignment in performance. The YOLOv8m student achieved an mAP of 79.4, deviating by only 1.3 points from the SAM 3 teacher (80.7). Furthermore, the $AP_{50}$ (93.6\%) and $AP_{75}$ (88.2\%) scores are remarkably similar to those of the teacher (93.6\% and 88.4\%, respectively). This similarity indicates that the lightweight models were able to learn the teacher's representations effectively, preserving not just the object localization (as shown by $AP_{50}$) but also the boundary precision reflected in the $AP_{75}$.

Table~\ref{tab:yolo_ap_metrics} highlights an important operational distinction: SAM 3's high latency (\approxm{1200} ms) restricts it to offline annotation, whereas the distilled YOLOv8s student operates at 6.10 ms—a \approxm{200}$\times$ speedup. This demonstrates that the proposed pipeline successfully compresses the foundation model's capabilities into a lightweight architecture suitable for real-time edge monitoring.

\subsection{Scenario-Stratified Evaluation}

Table~\ref{tab:custom_groups} summarizes the performance of the YOLOv8s student model across the eight scenarios defined in Section~\ref{sec:stratified-eval}.
By analyzing these distinct environments, we observe a clear correlation between scene complexity and model reliability.

% To move beyond aggregate metrics and identify specific failure modes of the proposed approach, we conducted a stratified analysis of the test set using the YOLOv8s student model. As seen in Figure \ref{fig:8-groups}, images were partitioned into eight distinct scenarios based on visual characteristics such as camera angle, pen layout, and lighting conditions.
% Groups 1 and 2 represent the Gestation phase. While similar, they differ in camera angle and field of view: Group 1 focuses on three stalls, whereas Group 2 captures a wider angle showing five.
% Group 3 is also part of the gestation phase but features a distinct pen layout with slatted flooring.
% Group 4 depicts the Nursery stage, filmed from a top-down perspective within a metal pen divided into two sections.
% Group 5 presents the most significant challenge, representing the Farrowing phase. These images contain sows alongside numerous piglets.
% The high density results in frequent severe occlusion, making accurate annotation and prediction particularly difficult.
% Group 6 shows pigs in the Growth phase, housed in a pen bisected by a central column.
% Notably, this group includes low-light night imagery, introducing luminosity-related detection difficulties.
% Group 7 captures the Estrus phase, characterized by a compact layout separated into three individual stalls.
% Finally, Group 8 also illustrates the Growth phase but utilizes an open pen layout without physical dividers.

\begin{table}[htbp]
\centering
\caption{Performance metrics per group (in \%)}
\label{tab:custom_groups}
\begin{tabular}{lcccccc}
\toprule
Group & Images & \textbf{$mAP$} & \textbf{$AP_{50}$} & \textbf{$AP_{75}$} & \textbf{$AP_{M}$} & \textbf{$AP_{L}$} \\
\midrule
Group 1 & 25 & 77.2 & 96.3 & 88.1 & - & 77.2 \\
Group 2 & 74 & 89.9 & 97.1 & 97.1 & - & 89.9 \\
Group 3 & 20 & 90.2 & 100.0 & 100.0 & - & 90.2 \\
Group 4 & 119 & 80.5 & 92.7 & 87.5 & 0.4 & 80.5 \\
Group 5 & 37 & 56.7 & 78.3 & 61.2 & 40.4 & 63.3 \\
Group 6 & 42 & 83.6 & 98.8 & 96.5 & 45.7 & 83.8 \\
Group 7 & 42 & 89.9 & 100.0 & 96.5 & - & 89.9 \\
Group 8 & 67 & 80.0 & 98.6 & 91.2 & 46.8 & 81.2 \\
\bottomrule
\end{tabular}
\end{table}

In scenarios characterized by minimal animal-to-animal occlusion (Groups~1, 2, 3, and~7), the student model achieves near-perfect detection rates, with $AP_{50}$ scores exceeding $99\%$ (reaching $100.0\%$ in Groups~3 and~7).
The physical separation in these views allows the foundation model (teacher) to generate unambiguous pseudo-labels, preventing the merging artifacts typically observed in crowded scenes. This indicates that, for monitoring setups characterized by low occlusion, foundation-model supervision presents a viable alternative to human ground truth, achieving comparable downstream performance.

A critical observation is found in Group 4, which constitutes the largest subset ($N=119$).
While the model performs robustly on large instances ($87.5\%$ AP), the accuracy for medium objects collapses to a misleading $0.4\%$.
Our manual audit revealed that this is not a failure of feature extraction, but a statistical artifact driven by class sparsity.
In this specific camera view, the zoom level renders the vast majority of pigs as ``Large'' ($>96^2$ pixels).
Consequently, the Ground Truth contains an almost negligible number of ``Medium'' instances—likely limited to partial pigs at the image periphery.
With such a small denominator, even a single missed detection or a minor segmentation discrepancy (e.g., a mask being a few pixels larger than the threshold) causes the metric to drop to near zero.
Thus, the low $AP_M$ reflects the absence of the class in the dataset rather than an inability to detect the animals.

\begin{figure}[h!]
  \centering
  \includegraphics[width=1.0\linewidth]{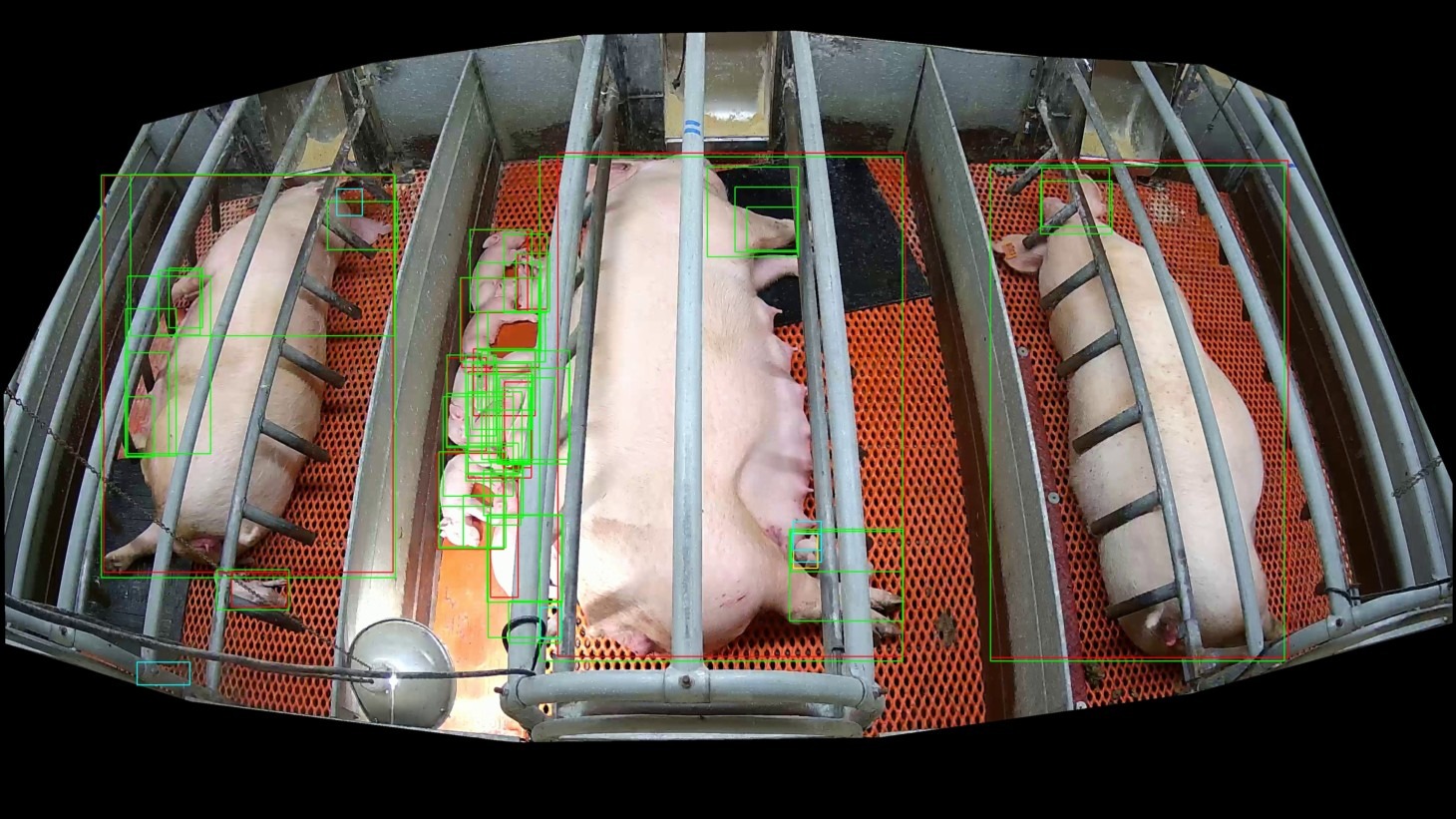}
  \caption{Captured from a top-down camera view, the image displays three sows housed in individual crates alongside their piglets.
Green bounding boxes indicate large objects, specifically identified sows and piglets.
Blue bounding boxes denote medium-sized objects, representing potential piglet detections or false positives.}
  \label{fig:group-five-example}
\end{figure}

Group 5 represents the true performance boundary of the proposed method, with mAP dropping to $56.7\%$.
Unlike the data artifact in Group 4, Group 5 exhibits a consistent degradation across all metrics, including a sharp decline in $AP_{75}$ to $61.2\%$.
As shown in Figure \ref{fig:group-five-example}, this scenario captures the complex visual environment of the Farrowing phase, where the significant size disparity between sows and piglets, combined with high-density clustering, challenges the foundation model's zero-shot capabilities.
The piglets often pile atop one another or are obscured by metal bars, leading to noisy pseudo-labels that propagate errors to the student model.

In summary, the experimental analysis indicates that the proposed distillation pipeline offers a favorable trade-off for precision livestock farming. While the stratified evaluation identified high-occlusion farrowing scenarios as a current boundary for zero-shot supervision, the system demonstrated robustness across the majority of production phases. The ability of the student model to maintain parity with the teacher's counting accuracy ($AP_{50}$), while reducing inference latency by \approxm{200}$\times$, reinforces the viability of compressing computationally heavy foundation models into lightweight detectors suitable for real-time edge deployment.

\subsection{Application Scenario}

Regarding the pervasive deployment of the developed model in real-world PLF scenarios, an edge computing architecture presents a promising approach. Recent benchmarking studies indicate that lightweight YOLOv8 variants are suitable for execution on embedded systems such as the NVIDIA Jetson Orin NX, particularly when leveraging hardware-specific optimizations like TensorRT \cite{b18}. Such embedded devices can accelerate deep learning inference to support real-time throughput while managing power constraints. Contextually, this localized processing architecture indicates that the reliance on constant, high-bandwidth cloud connectivity can be significantly reduced, or even rendered unnecessary in certain rural scenarios, thereby facilitating more robust continuous monitoring of the herd.

\section{Conclusions}

This study validates the efficacy of foundation model distillation as a scalable alternative to manual annotation for precision livestock monitoring. Our pipeline, which utilizes SAM 3 to supervise a lightweight YOLOv8 student, eliminates the bottleneck of human labeling while maintaining high downstream performance. Although aggregate metrics indicate a performance gap—specifically a 12.3\% decrease in mAP for the YOLOv8m variant compared to human supervision—this metric is heavily influenced by strict mask overlap criteria ($AP_{50-95}$) and scale definition artifacts in specific camera views. Crucially, the student model preserved a robust $AP_{50}$ of 93.6\%, confirming its reliability for the primary task of animal counting and localization.

The stratified scenario analysis revealed that the utility of this method is context-dependent. In environments characterized by spatial separation or individual stalls (Gestation and Estrus), the automated pipeline achieved competitive detection rates, proving functionally equivalent to human annotation. Conversely, the Farrowing phase (Group 5) was identified as the current boundary of zero-shot supervision, where extreme size disparity and dense piglet clustering degrade the teacher's signal.

Ultimately, the proposed SAM 3-to-YOLOv8 distillation offers a decisive computational advantage. By compressing the foundation model's knowledge (inference time \approxm{1200} ms) into a compact edge-deployable model (\approxm{6} ms), we achieved a \approxm{200$\times$} speedup with negligible loss in counting accuracy. For commercial deployments, we recommend this automated pipeline as a standard solution for Gestation and Nursery monitoring, while suggesting a hybrid ``human-in-the-loop'' approach for complex Farrowing environments where foundation models currently require refinement.

\subsection{Future Work}

To bridge the 12.3-point mAP gap and reach the human-annotated baseline, our future research will implement an active learning framework. We will utilize the student models trained on SAM 3 annotations to identify ``uncertain'' predictions—specifically instances where the model exhibits low confidence or high variance on the training data. By isolating these specific problematic frames for targeted manual annotation, we aim to maximize detection accuracy while minimizing the total human labor required. This hybrid strategy will evolve the current pipeline into a scalable, high-precision system suitable for diverse commercial pig production facilities.

%\todo[inline]{A formatação da bibliografia está inconsistente.}

\section*{Acknowledgment}

This work was supported by FAPES/UnAC (No. 1068/2023, P 2023-SGLQ7) through Sistema UniversidadES, and by CAPES/FAPES (No. 132/2021, P 2021-2S6CD) under the PDPG (Graduate Development Program - Strategic Partnerships in the States).

% The preferred spelling of the word ``acknowledgment'' in America is without 
% an ``e'' after the ``g''. Avoid the stilted expression ``one of us (R. B. 
% G.) thanks $\ldots$''. Instead, try ``R. B. G. thanks$\ldots$''. Put sponsor 
% acknowledgments in the unnumbered footnote on the first page.

\vspace{12pt}


\begin{thebibliography}{00}
\bibitem{b1} D. B. M. Yousefi, A. S. M. Rafie, S. A. R. Al-Haddad, and S. Azrad, ``A Systematic Literature Review on the Use of Deep Learning in Precision Livestock Detection and Localization Using Unmanned Aerial Vehicles,'' \textit{IEEE Access}, vol. 10, pp. 80071--80091, 2022, doi: 10.1109/ACCESS.2022.3194507.

\bibitem{b2} D. Berckmans, ``General introduction to precision livestock farming,'' \textit{Animal Frontiers}, vol. 7, no. 1, pp. 6--11, Jan. 2017, doi: 10.2527/af.2017.0102.

\bibitem{b3} T. Norton, C. Chen, M.L.V. Larsen, D. Berckmans,
Review: Precision livestock farming: building ‘digital representations’ to bring the animals closer to the farmer, Animal, Volume 13, Issue 12,
2019, Pages 3009-3017, ISSN 1751-7311, https://doi.org/10.1017 S175173111900199X.

\bibitem{b4} N. Carion, L. Gustafson, Y. T. Hu, S. Debnath, R. Hu, D. Suris, C. Ryali, K. V. Alwala, H. Khedr, A. Huang, and J. Lei, ``SAM 3: Segment anything with concepts,'' arXiv preprint arXiv:2511.16719, 2025.

\bibitem{b5} N. Ravi, V. Gabeur, Y. T. Hu, R. Hu, C. Ryali, T. Ma, H. Khedr, R. R\"adle, C. Rolland, L. Gustafson, and E. Mintun, ``SAM 2: Segment anything in images and videos,'' arXiv preprint arXiv:2408.00714, 2024.

\bibitem{b6} J. Seo, H. Ahn, D. Kim, S. Lee, Y. Chung, and D. Park, ``EmbeddedPigDet---Fast and Accurate Pig Detection for Embedded Board Implementations,'' \textit{Applied Sciences}, vol. 10, no. 8, Art. no. 2878, 2020, doi: 10.3390/app10082878.

\bibitem{b7} R. Gong, H. Zhang, G. Li, and J. He, ``Edge Computing-Enabled Smart Agriculture: Technical Architectures, Practical Evolution, and Bottleneck Breakthroughs,'' \textit{Sensors}, vol. 25, no. 17, Art. no. 5302, Aug. 2025, doi: 10.3390/s25175302.

\bibitem{b8} J. Redmon, S. Divvala, R. Girshick, and A. Farhadi, ``You only look once: Unified, real-time object detection,'' in \textit{Proc. IEEE Conf. Comput. Vis. Pattern Recognit. (CVPR)}, 2016, pp. 779--788.

\bibitem{b9} Li, J.; Ma, W.; Wei, Y.; Wang, T. Enhanced YOLOv8 for Robust Pig Detection and Counting in Complex Agricultural Environments. Animals
2025, 15, 2149. https://doi.org/10.3390/ani15142149

\bibitem{b10} T. Diwan, G. Anirudh, and J. V. Tembhurne, ``Object detection using YOLO: challenges, architectural successors, datasets and applications,'' \textit{Multimedia Tools and Applications}, vol. 82, pp. 9243--9275, 2023, doi: 10.1007/s11042-022-13644-y.

\bibitem{b11} Jiangong Li, Angela R. Green-Miller, Xiaodan Hu, Ana Lucic, M. R. Mahesh Mohan, Ryan N. Dilger, Isabella C.F.S. Condotta, Brian Aldridge, John M. Hart, Narendra Ahuja, ``Barriers to computer vision applications in pig production facilities,'' \textit{Computers and Electronics in Agriculture}, vol. 200, Art. no. 107227, 2022, doi: 10.1016/j.compag.2022.107227.

\bibitem{b12} M. V. Mendes Faria, T. Meireles Paixão, and F. de Assis Boldt, ``Comparative Evaluation of YOLO-family Detectors for Pig Detection in Precision Livestock Systems,'' RITA, vol. 33, no. 2, pp. 98–105, Mar. 2026, https://doi.org/10.22456/2175-2745.150940.

\bibitem{b13} J. Lee, H. Chae, S. Son, J. Seo, Y. Suh, J. Lee, Y. Chung, and D. Park, ``Sustainable Self-Training Pig Detection System with Augmented Single Labeled Target Data for Solving Domain Shift Problem,'' \textit{Sensors}, vol. 25, no. 11, Art. no. 3406, 2025, doi: 10.3390/s25113406.

\bibitem{b14} M. Wutke, C. Lensches, U. Hartmann, and I. Traulsen, ``Towards automatic farrowing monitoring---A Noisy Student approach for improving detection performance of newborn piglets,'' \textit{PLOS ONE}, vol. 19, no. 10, e0310818, 2024, doi: 10.1371/journal.pone.0310818.

% \bibitem{b13} A. M. Rickmann, S. L. Thorn, S. S. Ahn, S. Lee, S. Uman, T. Lysyy, R. Burns, N. Guerrera, F. G. Spinale, J. A. Burdick, and A. J. Sinusas, ``Using Foundation Models as Pseudo-label Generators for Pre-clinical 4D Cardiac CT Segmentation,'' in \textit{Proc. Int. Conf. Functional Imaging and Modeling of the Heart}, Cham, Switzerland: Springer Nature Switzerland, May 2025, pp. 253--265, doi: 10.48550/arXiv.2505.09564.

\bibitem{b15} A. M. Rickmann \textit{et al.}, ``Using Foundation Models as Pseudo-label Generators for Pre-clinical 4D Cardiac CT Segmentation,'' in \textit{Proc. Int. Conf. Functional Imaging and Modeling of the Heart (FIMH)}, vol. 15673, Cham: Springer, 2025, pp. 253--265, doi: 10.1007/978-3-031-94562-5\_23.

\bibitem{b16} C. Zeng, Y. Jiang, and A. Zhang, ``EfficientSAM3: Progressive Hierarchical Distillation for Video Concept Segmentation from SAM1, 2, and 3,'' arXiv preprint arXiv:2511.15833, 2025.

\bibitem{b17} Jiangong Li, Xiaodan Hu, Ana Lucic, Yiqi Wu, Isabella C.F.S. Condotta, Ryan N. Dilger, Narendra Ahuja, and Angela R. Green-Miller, ``Promote computer vision applications in pig farming scenarios: high-quality dataset, fundamental models, and comparable performance'' \textit{Journal of Integrative Agriculture}, 2024, doi: 10.1016/j.jia.2024.08.014.

\bibitem{b18} Aljami, H. M., Alrowais, N. A., AlAwajy, A. M., Alhrgan, S. O., Aldwaani, R. A., Alsawadi, M. S., Saqib, N. U., Alam, S. S., \& Alsubaie, R. (2026). Benchmarking YOLOv8 Variants for Object Detection Efficiency on Jetson Orin NX for Edge Computing Applications. Computers, 15(2), 74. https://doi.org/10.3390/computers15020074

\end{thebibliography}
\end{document}